\documentclass[conference]{IEEEtran}
\IEEEoverridecommandlockouts
\usepackage[inline]{enumitem}
\usepackage{algorithm}
\usepackage{algpseudocode}
\usepackage{amsmath}
\usepackage{color}
\usepackage{graphicx}
\usepackage{atbegshi}
\newcommand\numberthis{\addtocounter{equation}{1}\tag{\theequation}}
\usepackage{fancyhdr}
\fancypagestyle{firstpage}{
\lfoot[\footnotesize{\linewidth}]{\small{ \textcopyright 2019 IEEE. Personal use of this material is permitted. Permission from IEEE must be obtained for all other uses, in any current or future media, including reprinting/republishing this material for advertising or promotional purposes, creating new collective works, for resale or redistribution to servers or lists, or reuse of any copyrighted component of this work in other works.}}
}

\begin{document}

\title{Spiking Neural Network based Region Proposal Networks for Neuromorphic Vision Sensors}

\author{\IEEEauthorblockN{Jyotibdha Acharya}
\IEEEauthorblockA{HealthTech NTU\\Interdisciplinary Graduate School\\
Nanyang Technological University\\
Singapore}
\and
\IEEEauthorblockN{Vandana Padala}
\IEEEauthorblockA{School of Electrical and \\Electronic Engineering\\
Nanyang Technological University\\
Singapore}
\and
\IEEEauthorblockN{Arindam Basu}
\IEEEauthorblockA{School of Electrical and \\Electronic Engineering\\
Nanyang Technological University\\
Singapore}}

\maketitle

\thispagestyle{firstpage}
\setcounter{page}{1}
\begin{abstract}
This paper presents a three layer spiking neural network based region proposal network operating on data generated by neuromorphic vision sensors. The proposed architecture consists of refractory, convolution and clustering layers designed with bio-realistic leaky integrate and fire (LIF) neurons and synapses. The proposed algorithm is tested on traffic scene recordings from a DAVIS sensor setup. The performance of the region proposal network has been compared with event based mean shift algorithm and is found to be far superior ($\approx 50\%$ better) in recall for similar precision ($\approx 85\%$). Computational and memory complexity of the proposed method are also shown to be similar to that of event based mean shift. 
\end{abstract}

\IEEEpeerreviewmaketitle

\section{Introduction}

Asynchronous dynamic vision sensors are bio-inspired visual sensors that produce spikes corresponding to each pixel in their visual fields (also termed address-event representation or AER) where there is a change in light intensity \cite{yousefzadeh2018hybrid}. These sensors received significant attention from research community in recent years due to their distinct advantages over traditional frame based video cameras in terms of both power efficiency and memory requirement. Several hardware implementations of these AER sensors have been made in the past decade \cite{posch2014retinomorphic}\cite{berner2013240}. A number of new event based algorithms have been proposed in recent years to successfully process the data from these sensors\cite{basu2018low}. These algorithms are applied in various applications ranging from motion estimation \cite{orchard2015hfirst} and stereo-vision \cite{rogister2012asynchronous} to motor control \cite{delbruck2013robotic} and gesture recognition \cite{amir2017low}.

However, most of these event based algorithms are inspired by traditional computer vision algorithms and therefore, not particularly suitable for neuromorphic processing \cite{osswald2017spiking}. Biologically plausible spiking neural networks (SNN) have been shown to perform successfully in  complex tasks like image classification \cite{diehl2015unsupervised} and stereo vision \cite{osswald2017spiking}. Due to their unique asynchronous spike based data processing architecture, SNNs are inherently suitable for spiking input data.

With increasing demand in autonomous vehicles, smart surveillance and human-computer interaction etc, accurate real time object tracking has become a primary research area in computer vision community \cite{yilmaz2006object}. With the advent of CNN and deep learning, a number of deep learning based object tracking algorithms have been proposed \cite{ren2015faster} \cite{redmon2016you}. Most of these object tracking algorithms have two distinct phases:
\begin {enumerate*} [label=\itshape\alph*\upshape)]
\item region proposal and \item object classification. 
\end {enumerate*} 
While the region proposal network proposes multiple bounding boxes per frame where there might be an object, the object classification network runs on the proposed regions and predicts the class of the object. Recent object tracking algorithms have used selective search \cite{uijlings2013selective}, CNN based region proposal networks \cite{dai2016r} etc. for generating region proposals.

With the development of several low-power SNN processors (\cite{merolla2014million},\cite{davies2018loihi}), it is timely to revisit signal processing algorithms and recast them in terms of SNN building blocks. In this work, we propose a SNN based region proposal network (RPN) -- the first stage for most tracking algorithms \cite{dai2016r} and apply it to real recordings using an event based neuromorphic vision sensor (NVS) \cite{berner2013240}. While the benefit of NVS in foreground extraction for stationary cameras is well known, it has not been properly quantified to the best of our knowledge. We propose the first SNN based RPN as well as use standard tracking metrics of precision vs recall to evaluate the RPN operating on the NVS recording of traffic data.

\vspace{-0.2cm}
\section{ Materials and Methods}
\label{Materials and Methods}
\vspace{-0.1cm}
\subsection{Data Collection}
\vspace{-0.1cm}
AER based event data is acquired using a DAVIS sensor (resolution - 240 $\times$180) setup at a traffic junction. This setup captures the movement of various moving entities in the scene and the typical objects in the scene include humans, bikes, cars, vans, trucks and buses. Multiple recordings of varying duration are obtained at different distances and day/night settings and the comprehensive details of the recordings used in the paper are presented in Table . The sizes of various moving objects vary by an order of magnitude in any given scene (eg: Humans vs Buses)  and their velocities also span over a wide range (sub-pixel for humans to 5-6 pixels/frame for other fast moving vehicles) in the same recording. For reference purpose, typical size of a particular class of object (car in this case) across various recordings is provided in the fourth column in Table \ref{Dataset Description}. These recordings were manually annotated to generate the ground truth annotations of these objects in the scene.

\begin{table}
\centering
\caption{Dataset Details}
\begin{tabular}{ |p{1.5cm}||p{1.2cm}|p{1.2cm}|p{1.2cm}|p{1.2cm} | }

  \hline
  \multicolumn{5}{|c|}{Dataset Details} \\
 \hline
 Distance (m) & Lighting Condition & Duration (s) & Average Car Size & Number of Events\\
 \hline
 50   & Day  & 58.9898 & 40x20 &  927242 \\
 50    & Night  & 59.9599  & 38x18 &  771646\\
 100    & Day   & 60.0291 & 28x14 &  630885\\
 100   & Night  & 59.9599 & 27x14 &  480272 \\
 150    & Day  & 58.9897 & 19x11 &  583646\\
 150    & Night   & 59.9593 & 19x11 &  479242\\
 \hline
\end{tabular}
\label{Dataset Description}
\vspace{-0.4cm}
\end{table}

\subsection{Proposed Architecture}
The basic building blocks of our proposed SNN are leaky integrate and fire (LIF) neurons and synapses. The membrane potential ($V(t)$) is governed by the following differential equation:
\vspace{-0.5cm}
\begin{equation}
    \tau_m \frac{dV}{dt}=-(V(t)-V_{rest})+RI(t)
\end{equation}
where $V_{rest}$ is the rest potential, $I(t)$ is the total synaptic current, $R$ is the membrane resistance and $\tau_m $ is the membrane time constant. When the membrane potential reaches threshold voltage ($V_{th}$), the neuron fires (produces one output spike) and then resets to reset voltage ($V_{reset}$). After one spike, the neuron can not spike again within a refractory period ($t_{refractory}$). The synapses are modelled using  exponentially decaying EPSCs i.e., when a spike arrives, the conductance ($g$) of the synapse increases instantaneously and decays otherwise according to:
\vspace{-0.3cm}
\begin{equation}
    \tau_g \frac{dg}{dt}=-g
\end{equation}
where $\tau_g$ is the synaptic time constant. All the neurons and synapses in a layer have the same neuron and synaptic parameters.

Our proposed architecture is a three layered network with first two asynchronous event driven layers and one final clustering layer that converts the event based outputs of previous layer to frame based outputs for visualization and evaluation. The architecture is as follows:
\subsubsection{Refractory Layer}
The size of the refractory layer is same as the input image ($H\times L$) and each input is connected to one neuron in the refractory layer in 1:1 connections. The neurons in this layer have large refractory period and small threshold voltage. The importance of this layer is two-fold. Firstly, the output of DVS sensors often contains significant amount of noise \cite{padala2018noise} and the poor SNR affects the effectiveness of further processing of the events. With proper tuning of the refractory period, a sizeable fraction of these noisy events are eliminated without any considerable loss of signal and thereby, SNR is improved. As a result, we get significantly smoother tracking boxes in further layers. Secondly, as this layer filters off a large fraction of the input events, the computational complexities of the further event based layers are considerably reduced.
\subsubsection{Convolution Layer}
The convolution layer operates on the spikes produced by the refractory layer. Each neuron in the convolution layer is connected to a sliding square window of size $W\times W$ and stride $S$. Each convolutional neuron spikes in response to increased activity within its corresponding window and produces one region proposal box of size $W\times W$. These neurons have no refractory period (since the input to this layer is already sparse enough) and have relatively larger threshold to handle larger synaptic currents coming from multiple input neurons.
\subsubsection{Local Excitation}
In a variant of the model, we have also proposed recurrent lateral connections in the convolution layer where each neuron is connected to its immediate neighbors through excitatory connections. 
\subsubsection{Clustering Layer}
This is the only frame based layer of our proposed architecture. In this layer, all the region proposal boxes generated by the convolution layer withing a given frame duration are accumulated and all the neighboring boxes are clustered together to form larger boxes. Since the convolution layer only produces fixed size region proposal boxes, this layer is necessary to combine the boxes to actual shapes of objects.

\begin{figure}[!t]
\centering
\includegraphics[width=\linewidth]{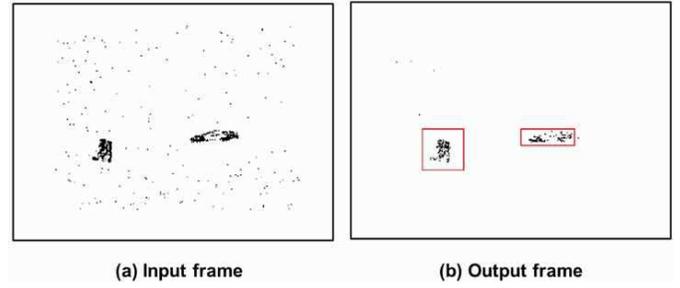}
\caption{Visualization of RPN input and output: input frame shows a scene with one car and two humans (a) and the corresponding output frame shows the region proposals in red (b). The denoising in the output frame is done by the refractory layer while the region proposal is done by convolution layer and clustering layer.}
\label{layer_visualization}
\vspace{-0.6cm}
\end{figure}

The proposed algorithm is summarized in algorithm \ref{algo}. Figure \ref{layer_visualization} shows a sample frame of the input data and corresponding output frame.

\begin{algorithm}
\caption{SNN based RPN}
\label{algo}
\begin{algorithmic}[1]
\scriptsize
\For{each input event}
\\modify refractory layer membrane voltages
\\feed-forward any refractory spike produced
\EndFor
\For{each refractory spike}
\\modify refractory synaptic currents
\\modify convolution layer membrane voltages
\\modify recurrent synaptic currents (optional)
\\feed-forward any region proposal produced
\EndFor

\For{each frame}
\\accumulate all region proposal boxes
\For{each region proposal box}
\If{There is an adjacent region proposal box}
 \\ \hspace{\algorithmicindent} \ \ combine both boxes to produce larger region proposals
\EndIf
\EndFor
\EndFor
\end{algorithmic}
\end{algorithm}
\vspace{-0.4cm}

\subsection{Computational Complexity \& Memory}
\label{Computational Complexity}

For variable updates during each event, we have used Euler method \cite{skocik2014capabilities} for calculating event based parameter updates. For the refractory layer, for each neuron, we need to store the membrane potential and the timestamp when the neuron received the last input spike. When a new spike comes at a input neuron, 5 operations are required to update the corresponding refractory membrane potential. So, if a b bit number is used to store each variable, refractory layer requires $2\times H\times L\times b$ bit memory and 5 operation per event.

Similarly, in the convolution layer, $5\times W\times W$ operations are required to update each synaptic conductance corresponding to one convolution layer neuron, $W\times W-1$ additional operations to compute the total input current and $5$ operations to update the membrane voltage. So, in total, $6\times W\times W+4$ operations are required to update each convolution layer neuron. Now, when there is a refractory spike, the membrane voltage of the convolutional neurons connected to that neuron will be updated. Since we are using a square window with $<50\%$ overlap, one refractory neuron is connected to at most 4 convolution neurons. So, a total of maximum $24\times W\times W+16$ operations are required per refractory spike. Now, as for memory requirement of this layer, $2\times H\times L\times b$ bits are required to store the synaptic conductances and corresponding last spike times. An additional $2\times M\times N\times b$ bits are required to store membrane potentials and corresponding last spike times for a convolution layer of dimension $M\times N$. 

Finally,  For the clustering layer, for each frame, all the region proposal boxes are iterated over and the neighboring boxes are combined. So, for r region proposals, $r\times(r-1)$ operations are required to compare box locations. The clustering layer will require a buffer of size $2\times r\times  b$ bits (2 numbers representing each proposed box) to accumulate the region proposals generated during a frame.

So, for total input spikes $K_{inp}$, total refractory layer spikes $\alpha\times K_{inp}$ ($\alpha << 1$ since refractory layer produces significantly less spikes at output) and total frames $F$, the total number of computes for tracking a recording without lateral excitation is given by:
\vspace{-0.3cm}
\begin{align*}
    C_{total} = & K_{inp}\times 5 +\alpha\times K_{inp}\times(24\times W\times W+16) \\
    & +F\times r\times(r-1) \numberthis
\end{align*}
\vspace{-0.2cm}
The total memory required in bits is given by:
\begin{align*}
\label{eq4}
    B_{total} = & 2\times H\times L\times b + 2\times H\times L\times b + \\
    & 2\times M\times N\times b + 2\times r\times b \numberthis
\end{align*}

\subsection{Evaluation Metrics}
To evaluate our region proposal network, we have adopted a precision vs recall curve based metric, traditionally used by computer vision community. The metric used to quantify the quality of proposed boxes is Intersection over Union ratio (IoU) defined as:
\begin{align*}
    IoU=\frac{\text{intersection of proposed \& ground truth box}}{\text{union of proposed \& ground truth box}}
\end{align*}
A certain threshold is defined based on IoU (e.g. IoU 0.5). Proposal boxes with IoU values larger than that threshold value are considered correct region detection (true positive box). Then, the performance of the tracker is evaluated on precision (true positive boxes/total proposal boxes) and recall (true positive boxes/total ground truth boxes) calculated over all the frames of the video. Parameter variation of the architecture produces different precision and recall values and therefore, the precision vs recall curve represent the performance of the region proposal algorithm in its entirety.

Although, IoU is perfectly suitable to evaluate region proposals for end to end object tracking, in case of standalone region proposal networks like the one we described, having proposal boxes larger than ground truth boxes is more advantageous than having proposal boxes smaller than ground truth boxes since larger boxes will ensure no loss of information to the object classifier and the classifier can be trained to tighten the proposal box if required. Since IoU is symmetric with respect to both ground truth and proposed boxes, it does not capture this distinction. So, we proposed another metric, fitness score, to evaluate the fitness of region proposal.
\begin{align*}
    \text{Fitness Score} = \frac{\text{intersection of proposed \& ground truth box}}{\text{area of ground truth box}}
\end{align*}

\section{Results and Discussions}
\label{Results and Discussions}
\begin{figure}[!ht]
\centering
\includegraphics[width=\linewidth]{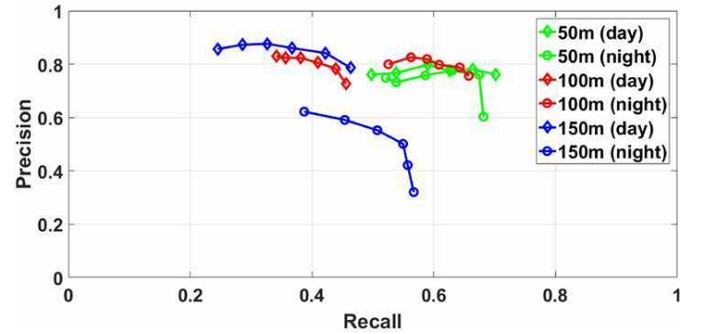}
\caption{Precision-recall curve for six recordings: sensor-object distance shows significant impact on the recall value.}
\label{first_figure}
\vspace{-0.1cm}
\end{figure}

For our first experiment, we have obtained the precision-recall curves for all 6 recordings in our dataset by varying the spiking threshold of convolution layer neurons (Fig. \ref{first_figure}). All the curves are plotted for a fixed IoU 0.3. Lower thresholds result in lower precision and higher recall while higher thresholds result in higher precision and lower recall. The recall value is higher for both 50 m day and night. This is the result of a smaller sensor-object distance. 150 m night curve shows significantly smaller precision due to unfavourable light condition (night) and much larger sensor-object distance.

\begin{figure}[!ht]
\centering
\includegraphics[width=\linewidth]{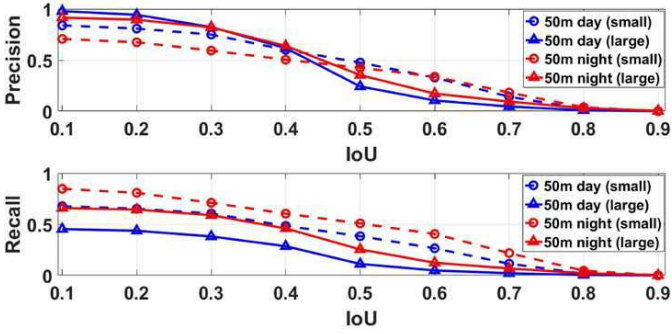}
\caption{IoU curve: smaller window size results in more accurate region proposals as evident from higher precision and recall for higher IoU values.}
\label{second_figure}
\vspace{-0.5cm}
\end{figure}

Now, in our proposed algorithm, the clustering is done entirely on the convolution layer proposals instead of original pixels. Although this ensures a significant saving in computational complexity, the resolution of the final boxes are limited by the dimension of convolution layer boxes. To explore the performance-complexity trade-off, we varied the IoU for fixed threshold value with different window sizes. Fig. \ref{second_figure} shows the precision and recall as a function of IoU for two different recordings with two different window sizes. Smaller window size results in a overall higher recall and better precision at higher IoU values. This goes to show that smaller window sizes produce more accurate region proposals due to availability of better resolution at the cost of higher computational complexity.

\begin{figure}[!ht]
\centering
\includegraphics[width=\linewidth]{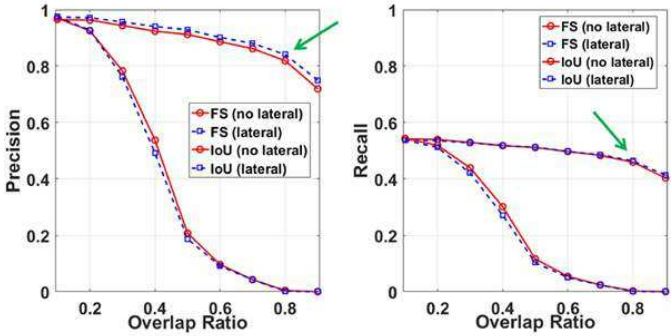}
\caption{Lateral excitation: precision and recall curve for 100 m (day) measured using IoU and fitness score (FS). Lateral excitation shows better precision at higher overlap ratios for FS measurement. For overlap ratio 0.8, lateral excitation improves precision by $2\%$ without loss of recall (marked by arrow).}
\label{forth_figure}
\vspace{-0.3cm}
\end{figure}

To examine the effect of lateral excitatory connections, we plotted the precision and recall curves for the same recording (100 m day) with and without lateral connections. We have seen in fig. \ref{second_figure} that IoU is not particularly suitable for measuring the performance of our architecture for higher IoU values due to resolution limitation of the algorithm. So, we evaluated the curves here using both IoU and fitness score (FS) parameters (Fig.  \ref{forth_figure}). While for IoU, lateral connections does not seem to yield any improvement, for FS, lateral connections show performance improvement over base architecture. While the recall remains similar with or without lateral connections, precision of the RPN shows improvement for lateral excitation case. Quantifying the effect of recurrent lateral connections, both excitatory and inhibitory, will require further investigation.

\begin{figure}[!ht]
\centering
\includegraphics[width=\linewidth]{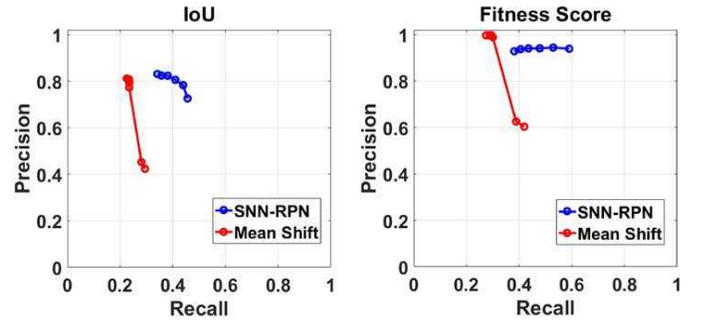}
\caption{Comparison with event based mean shift algorithm: precision-recall curve for 100 m (day) measured using IoU and fitness score. SNN-RPN outperforms mean shift for IoU based measurements while mean shift obtains slightly higher precision for fitness score based measurement at significantly smaller recall.}
\vspace{-0.3cm}
\label{fifth_figure}
\end{figure}

To benchmark our algorithm, we have compared it with the event based mean shift algorithm described in Delbruck et. al (2013) \cite{delbruck2013robotic}. For a fair comparison, we applied the mean shift algorithm on the de-noised data at the output of the refractory layer. Fig. \ref{fifth_figure} shows the precision-recall curves (measured using both IoU and fitness score) for both algorithms on 100 m (day) recording. While our algorithm shows clear advantage both in terms of precision and recall for IoU based measurement, for fitness score based measurement, mean shift achieves slightly higher precision at the cost of significantly reduced recall value. Moreover, our proposed algorithm generates a much stable ROC curve which signifies lesser dependence on fine tuning the threshold parameter.

Finally, based on the formula developed in section \ref{Computational Complexity}, we calculate the total number of operations and memory requirements for the algorithm. For the recording in our dataset the value of $\alpha$ ranges from $0.1-0.2$ and mean value of $r$ ranges from $4-6$. So, assuming $\alpha=0.15$ and $r=5$, for sensor input dimension $180\times 240$, convolution layer size $15\times 20$, window size $16\times 16$ and a frame rate of $30$ FPS the architecture requires $0.9$ Kops/event. For reference, if the convolution and clustering layer is replaced by mean shift tracker with similar configurations, it will require approximately $1.2$ Kops/event. Now, if the variables are saved as 8-bit numbers, i.e. $b=8$, the total memory required is given by $1.3$ Mbits. It can be clearly seen from equation \ref{eq4} that the total memory requirement is dominated by the memory required by noise filtering layer (refractory layer). For the mean shift tracker, the required memory will be approximately $1.05$ Mbits. So, the proposed algorithm performs significantly better than event based mean shift for similar memory requirement and computational cost.
\vspace{-0.4cm}
\section{Conclusion}
\label{Conclusion}
In this work, we have proposed a three-layer SNN based region proposal network for event based processing of neuromorphic vision sensor recordings of traffic scenes. We have also introduced evaluation metrics for the region proposal network analogous to traditional computer vision techniques. The proposed algorithm is tested for different sensor-object distance and light conditions (day/night). The precision-recall trade-off is parameterized by neuron firing threshold. The resolution of the proposed boxes and thereby their accuracy is dependent on convolution window size and therefore, there is also an apparent computation-performance trade-off.
Although this work is limited to only region proposals, we plan to extend this work in future to include a classification layer to evaluate its performance more accurately. We also want to combine a SNN based classifier similar to Diehl et. al. \cite{diehl2015unsupervised} with our architecture to design an end to end SNN based object detection framework.

\vspace{-0.3cm}

\section{Acknowledgment}
The authors would like to thank Dr. Garrick Orchard (National University of Singapore) for providing access to the traffic recordings data.

\bibliographystyle{IEEEtran}
\bibliography{example.bib}

\end{document}